\pdfoutput=1

\documentclass[11pt]{article}

\usepackage{acl}

\usepackage{times}
\usepackage{latexsym}

\usepackage[T1]{fontenc}

\usepackage[utf8]{inputenc}

\usepackage{microtype}

\usepackage{inconsolata}

\usepackage{bm}
\usepackage{array}
\usepackage{xspace}
\usepackage{amssymb}
\usepackage{amsmath}
\usepackage{enumitem}
\usepackage{booktabs}
\usepackage{graphicx}
\usepackage{multirow}
\usepackage{multicol}
\usepackage{makecell}
\usepackage{listings}
\usepackage{subfigure}
\usepackage{tablefootnote}
\usepackage[normalem]{ulem}
\usepackage[ruled]{algorithm2e}

\setlist[itemize]{leftmargin=5mm, itemsep=0mm}

\definecolor{comment}{rgb}{0,0.6,0}
\definecolor{key}{rgb}{0.84, 0.44, 0.84}
\definecolor{func}{rgb}{0.16, 0.72, 0.86}
\definecolor{var}{rgb}{1, 0.33, 0.33}
\definecolor{true}{rgb}{1, 0.78, 0.02}

\newcommand{\model}{FIM-SE\xspace}
\newcommand{\ie}{\emph{i.e.,}\xspace}

\newcommand{\paratitle}[1]{\vspace{1ex}\noindent{\bf #1}}

\title{Empowering Character-level Text Infilling by Eliminating Sub-Tokens}

\author{
  Houxing Ren$^{1,3}$ \quad Mingjie Zhan$^2\footnotemark[1]$ \quad Zhongyuan Wu$^2$ \quad Hongsheng Li$^{3,4,5}$\thanks{Corresponding author.} \\
  $^1$Shanghai Jiao Tong University \quad $^2$SenseTime Research \\
  $^3$Shanghai Artificial Intelligence Laboratory \quad $^4$CUHK MMLab \quad $^5$CPII under InnoHK \\
  \{renhouxing,georgewzy01\}@gmail.com \\
  zhanmingjie@sensetime.com \quad hsli@ee.cuhk.edu.hk \\
  \url{https://raccoon.sensetime.com/code}
}

\begin{document}
\maketitle
\begin{abstract}
In infilling tasks, sub-tokens, representing instances where a complete token is segmented into two parts, often emerge at the boundaries of prefixes, middles, and suffixes. Traditional methods focused on training models at the token level, leading to sub-optimal performance in character-level infilling tasks during the inference stage. Alternately, some approaches considered character-level infilling, but they relied on predicting sub-tokens in inference, yet this strategy diminished ability in character-level infilling tasks due to the large perplexity of the model on sub-tokens. In this paper, we introduce FIM-SE, which stands for Fill-In-the-Middle with both Starting and Ending character constraints. The proposed method addresses character-level infilling tasks by utilizing a line-level format to avoid predicting any sub-token in inference. In addition, we incorporate two special tokens to signify the rest of the incomplete lines, thereby enhancing generation guidance. Extensive experiments demonstrate that our proposed approach surpasses previous methods, offering a significant advantage. Code is available at \url{https://github.com/SenseLLM/FIM-SE}.
\end{abstract}

\section{Introduction} \label{sec:intro}

The Transformer~\cite{transformer2017ashish} decoder-only architecture has proven highly effective in various natural language processing~(NLP) tasks. 
This success has paved the way for the development of advanced causal decoder-only models like GPT-4~\cite{gpt42023openai}, PaLM~\cite{palm2023,palm22023}, Llama~\cite{llama2023hugo,llama22023hugo,codellama2023roziere}, and Falcon~\cite{falcon2023guilherme}. 
These innovative models excel at generating coherent and contextually relevant responses to natural language prompts, showcasing state-of-the-art performance across various tasks, including question answering~\cite{rag2020lewis}, logical reasoning~\cite{cot2023kojima}, and code synthesis~\cite{starcoder2023li,codellama2023roziere}. 

\begin{table}[t] \small
\centering
\caption{Examples for random splitting with Llama tokenizer, where the red, blue, and green text indicates the prefix, the middle, and the suffix, respectively. These four rows represent the pieces after randomly splitting, the sentence after exchanging suffix and middle, tokenized results, and token IDs, respectively.}
\subtable[The first splitting case.]{
    \begin{tabular}{c|l} \toprule
        Pieces & \color{red}{A f}\color{blue}{ine }\color{green}{day.} \\ \midrule
        Reorder & \color{red}{A f}~\color{green}{day.}~\color{blue}{ine } \\ \midrule
        Tokens & [`A', `\_f', `day', `.', `ine', `\_'] \\ \midrule
        IDs & [29909, 285, 3250, 29889, 457, 29871] \\ \bottomrule 
    \end{tabular}
}
\subtable[The second splitting case.]{
    \begin{tabular}{c|l} \toprule
        Pieces & \color{red}{A fi}\color{blue}{ne }\color{green}{day.} \\ \midrule
        Reorder & \color{red}{A fi}~\color{green}{day.}~\color{blue}{ne } \\ \midrule
        Tokens & [`A', `\_fi', `day', `.', `ne', `\_'] \\ \midrule
        IDs & [29909, 5713, 3250, 29889, 484, 29871] \\ \bottomrule 
    \end{tabular}
}
\label{tab:intro_example}
\end{table}

Despite the success, the proficiency of these models is somewhat limited in tasks involving text infilling, which aims to generate text at a specific location within a prompt, while conditioning on both a prefix and a suffix.
The main reason is their intrinsic left-to-right autoregressive design. 
To address this issue, CM3~\cite{cm32022armen} introduced the causal masking objective, placing a mask token at the intended fill location and completing the fill at the end. 
In contrast, FIM~\cite{fim2023mohammad} proposed a fill-in-the-middle technique, which randomly divides documents into three segments and tags them with three special tokens. 
This technique then rearranges the middle and suffix segments, to use the prefix and the suffix to predict the middle segment in auto-regressive format. 
With these methods, decoder-only based models can effectively handle various infilling tasks and achieve excellent performance. 

\begin{figure}[t]
    \centering
    \subfigure[Experiment.]{ \label{fig:intro_exp}
        \includegraphics[width=0.45\columnwidth]{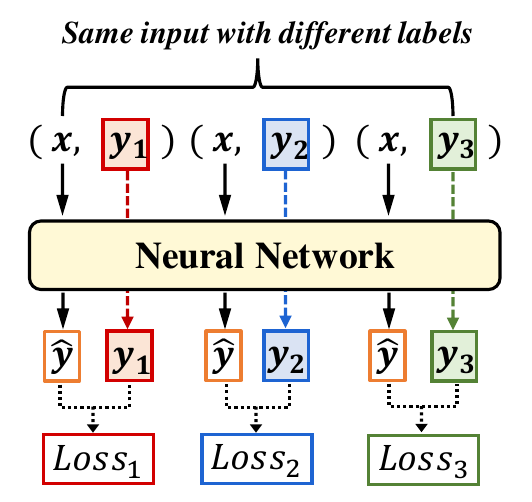}
    }
    \subfigure[Probabilities.]{ \label{fig:intro}
        \includegraphics[width=0.45\columnwidth]{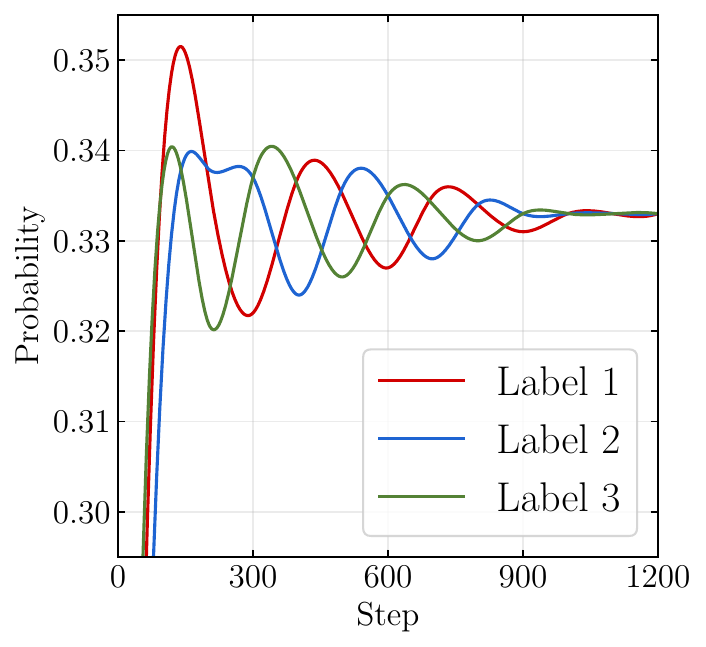}
    }
    \caption{The probabilities of prediction when inconsistent labels appear in the training data.}
    
\end{figure}

However, employing the aforementioned methods may introduce inconsistencies during training.
This arises from the potential division of a single token into multiple sub-tokens, as exemplified in Table~\ref{tab:intro_example}.
As we can see, due to character-level random splitting, the same prefixes~([29909]) have inconsistent objectives~(285 and 5713, respectively) in different cases.
The inconsistent objectives will significantly impact the model's perplexity, especially on sub-tokens.
To illustrate, we construct a simple experiment on a classification task shown in Figure~\ref{fig:intro_exp}. 
The training data only contains three samples and they have the same input but different labels.
We train a simple network on the training data and record the predicted probabilities of the three labels at each training step.
As shown in Figure~\ref{fig:intro}, the predicted probabilities for these three classes converge to 0.33, indicating a large perplexity of the model on the inconsistent objectives.
The large perplexity on sub-tokens makes the probability of error increase when predicting a sub-token.
This phenomenon is notably detrimental in sensitive tasks such as code completion, where even a minor error in any token can result in program malfunction.
As a result, previous approaches have yet to fully inspire the potential of Transformer decoder-only models in infilling tasks.

To effectively address the issue, it is crucial to acknowledge and resolve an inherent conflict. 
(1) We need to avoid the model predicting sub-tokens. In the infilling training mode, the model's perplexity in sub-tokens is large, resulting in the low accuracy of predicting sub-tokens.
(2) We need to output a sub-token when the user only writes part of a token. Because it is necessary to ensure that the output fits the context. If we directly drop several tokens to make sure no sub-tokens exist, the model's output may no longer align with the removed context, rendering it unreasonable in practical use.

Based on these concerns, we propose \model, which stands for Fill-In-the-Middle with both Starting and Ending character constraints. 
Our method enhances the organizational framework of FIM~\cite{fim2023mohammad} to concurrently address the two scenarios mentioned earlier.
In simple terms, we transfer the random-span infilling task to the multi-line infilling task. 
Specifically, after random character level splitting, we utilize two distinct special tokens to mark the \emph{Last line of the Prefix~(L-Prefix)} and the \emph{First line of the Suffix~(F-Suffix)}. 
The model is then tasked to generate text at line level that starts with \emph{L-Prefix} and ends with \emph{F-Suffix}.
Their inclusion in the prompt simplifies the task for the model, facilitating the generation of text that seamlessly starts with \emph{L-Prefix} and ends with \emph{F-Suffix}.
Overall, this method is designed to unlock the capabilities of decoder-only models in infilling tasks.

The core contribution of the paper is that we design a novel training method for the infilling task, a solution designed to effectively mitigate conflicts mentioned above in infilling tasks. Our method can effectively eliminate any potential inconsistencies and earnestly guarantee that the model's output aligns cohesively with the given context. Extensive experiments demonstrate the effectiveness of the proposed method on infilling tasks while not compromising code generation capabilities. Based on Code Llama 13B, our approach not only achieves an 8.8\% enhancement in the Humaneval random-span infilling task, with substantial improvements of 11.5\% and 10.7\% in the single-line and multi-line infilling tasks respectively, but also maintains minimal impact on the model's performance in code generation tasks.
\section{Related Work} \label{sec:rel}

\subsection{Large Language Models for Infilling}

Various Large Language Models (LLMs) have been developed for general generation tasks~\cite{llama2023hugo,llamalong2023wenhan,palm2023,falcon2023guilherme,mistral2023albert,baichuan2023aiyuan,qwen2023jinze,Claude2023Anthropic,Mixtral2023ABS240104088,Gemini2023ABS231211805}. Most of these models adopt a left-to-right autoregressive generation approach due to its effectiveness, as validated by research such as GPT~\cite{gpt12018radford,gpt22019radford,gpt32020tom,instructgpt2022long,gpt42023openai}. In the realm of code-related tasks, where infilling is essential, LLMs are specifically trained for this task. For instance, InCoder~\cite{incoder2023daniel} utilizes a causal masking objective, while SantaCoder~\cite{santacoder2023loubna},  StarCoder~\cite{starcoder2023li}, and Code Llama~\cite{codellama2023roziere}  employ the fill-in-the-middle technique introduced by FIM~\cite{fim2023mohammad}.

\subsection{Text Infilling Models}

The infilling task plays a crucial role in numerous real-world applications, including document editing\footnote{\url{https://copilot.microsoft.com}} and code completion\footnote{\url{https://github.com/features/copilot}}.
Three common Transformer~\cite{transformer2017ashish} architectures are capable of executing this task, \ie encoder-only, encoder-decoder, decoder-only.
In encoder-only architectures, masked language modeling is employed as the pre-training task, exemplified by models like BERT~\cite{bert2019jacob} and RoBERTa~\cite{roberta2019yinhan}. These models are designed to infill brief spans, ranging from a single token~\cite{bert2019jacob} to a word~\cite{wwm2021yiming}, and even several contiguous tokens~\cite{spanbert2020mandar}.
In encoder-decoder architectures, a common approach involves masking several tokens in the encoder and then tasking the model with decoding the complete sentence, as exemplified by MASS~\cite{mass2019kaitao}. Additionally, models like BART~\cite{bart2020mike} and T5~\cite{t52020colin} have introduced an infilling noising method. This technique replaces multiple tokens with a single mask token, challenging the model to decode the masked span.
In decoder-only architectures, several methods are employed for infilling tasks. The Insertion Transformer~\cite{insertion2019mitchell} instructs the model to first determine the location for the next token, followed by the token prediction itself. Meanwhile, GLM~\cite{glm2022zhengxiao}, CM3~\cite{cm32022armen}, and InCoder~\cite{incoder2023daniel} adopt a different approach. They shift the target span to the end of the context, employing left-to-right autoregressive modeling for training. In addition, MIM~\cite{MIM2023NguyenKC23} proposed to use both forward and backward LMs that share parameters to significantly enhance the performance.

Most of these models are designed for token-level infilling tasks, which often don't align with real-world applications due to the incomplete nature of the final token in actual prompts. FIM~\cite{fim2023mohammad} explored various levels of spans, \ie line level, token level, and character level. As shown in their results, models trained with line-level or token-level spans perform poorly on character-level infilling tasks. 
To enhance the performance of models trained on token-level spans in character-level infilling tasks, token healing was proposed to fix tokenization artifacts that normally arise at the boundary between the end of a prompt and the beginning of a set of generated tokens\footnote{\url{https://github.com/guidance-ai/guidance/blob/main/notebooks/tutorials/token_healing.ipynb}}.
While it effectively bridges the gap between the prefix and generated text, it falls short in handling the transition between generated text and the suffix, highlighting the need for further research in character-level infilling.
\section{Preliminaries} 

In this section, we provide a straightforward introduction to the FIM~\cite{fim2023mohammad} method and conduct a theoretical analysis of how inconsistent labeling affects the model's perplexity.

\subsection{Fill-In-the-Middle~(FIM)} \label{sec:pre_fim}

FIM is designed to train models to complete the central sections of documents. 
This approach involves joint training on a combined dataset of traditional left-to-right sequences and data transformed by FIM, with an infilling rate reaching as high as 90\%. 
According to experimental results, FIM maintains the autoregressive test losses of the left-to-right models without incurring significant costs, and it only slightly impacts the performance in downstream evaluations~\cite{santacoder2023loubna}.

In a particular document, FIM segments a document into three distinct parts: the prefix, the middle, and the suffix. It introduces three levels of segmentation: single-line, multi-line, and random-span.
Because random-span is more in line with actual usage conditions, previous studies~\cite{codellama2023roziere,starcoder2023li} usually trained the model with random-span level splitting. After splitting, it moves the middle piece to the end as
\begin{equation*}
    \text{doc} \to \text{(pre, mid, suf)} \to \text{(pre, suf, mid)},
\end{equation*}
then concatenate the three pieces using special tokens as
\begin{equation*}
    \text{<PRE> pre <SUF> suf <MID> mid <EOT>}.
\end{equation*}
This mode is termed Prefix-Suffix-Middle (PSM) mode. Additionally, FIM introduced the Suffix-Prefix-Middle (SPM) mode, which interchanges the positions of the prefix and suffix. A variant of the SPM mode is also proposed, maintaining the same structure as the PSM mode. Detailed descriptions of these modes are provided in Appendix~\ref{appendix:fim}.

\subsection{Impact of Inconsistent Labels} \label{sec:met_impact}

When the FIM method employs the random-span approach, a training sample can contain up to four sub-tokens. This can potentially lead to inconsistent labels that indicate the same input but with different labels. This issue becomes particularly critical when the model is required to predict a sub-token. In Section~\ref{sec:intro}, we construct a simple experiment to illustrate that this inconsistency can significantly affect the model's perplexity. Here, we offer a theoretical analysis to further elucidate this phenomenon.

We are considering a classification task involving $n$ classes, where each sample's label is associated with one of $m$ different categories across various training instances. Let $\bm{y} \in \{0, 1\}^{n}$ represent the actual label, and $\hat{\bm{y}} \in \mathbb{R}^{n}$ represent the predicted probabilities. In this context, the cross-entropy loss is computed as
\begin{equation}
    \mathcal{L} = - \sum_{i = 0}^{n} \bm{y}_i \log \hat{\bm{y}}_i.
\end{equation}
Assuming that the first $m$ elements of $\bm{y}$ (i.e., $\bm{y}_1, \dots, \bm{y}_m$) are set to 1, while the rest are 0, the loss function is defined as
\begin{equation}
    \mathcal{L}(\hat{\bm{y}}) = - \sum_{i = 0}^{m} \log \hat{\bm{y}}_i.
\end{equation}
Then our objective can be expressed as
\begin{equation} \label{eq:condition}
    \hat{\bm{y}}^{*} = \arg\max \mathcal{L}(\hat{\bm{y}}),~~~~s.t. \sum_{i = 0}^{n} \hat{\bm{y}}_i = 1.
\end{equation}
Here, we introduce the concept of the Lagrange Multiplier, which enables us to formulate the target function as
\begin{equation}
    \mathcal{L}(\hat{\bm{y}}, \lambda) = - \sum_{i = 0}^{m} \log \hat{\bm{y}}_i - \lambda(\sum_{i = 0}^{n} \hat{\bm{y}}_i - 1).
\end{equation}
We then calculate the partial derivatives of $\hat{\bm{y}}_1, \dots, \hat{\bm{y}}_m$ respectively, which allows us to obtain
\begin{equation}
    \frac{\partial \mathcal{L}(\hat{\bm{y}}, \lambda)}{\partial \hat{\bm{y}}_i} = - \frac{1}{\hat{\bm{y}}_i} - \lambda ~~~\text{when}~~~ i = 1, \dots, m. 
\end{equation}
By setting these derivatives to zero, we obtain
\begin{equation}
    \hat{\bm{y}}_1^{*} = \dots = \hat{\bm{y}}_m^{*} = - \frac{1}{\lambda}.
\end{equation}
Since $\hat{\bm{y}}_{m+1}, \dots, \hat{\bm{y}}_{n}$ are not included in the objective function of Eq.~\eqref{eq:condition}, and given that the logarithm is a monotonically increasing function, setting these values to 0 would maximize the objective function. Consequently, the condition is formulated as $\sum_{i = 1}^{m} \hat{\bm{y}}_i = 1$. By incorporating this condition, we derive
\begin{equation}
    \hat{\bm{y}}_1^{*} = \dots = \hat{\bm{y}}_m^{*} =  \frac{1}{m}.
\end{equation}
We have now completed the proof, demonstrating that when a data point is labeled differently across various samples, the model tends to assign an equal probability of $\frac{1}{m}$ to each label. This behavior leads to a large perplexity of the model, which further suggests its limited modeling capability. \

We consider this proof to be a microscopic explanation, which demonstrates that the perplexity of sub-tokens will be higher. In contrast, macroscopically speaking, we assume that the probability of the next token prediction obeys a certain distribution. Then, these sub-tokens are outliers. The presence of several outliers in each piece of training data will result in a low confidence in the model, that is, a high degree of perplexity. 

This phenomenon is notably detrimental in sensitive tasks. For example, the initially predicted token in practical completion is usually a sub-token. The higher perplexity of the first token has little impact on the overall quality of the generated text, but in some sensitive tasks such as code completion, even a minor error in any token can result in program malfunction.

\begin{figure*}[t]
    \centering
    \includegraphics[width=0.95\textwidth]{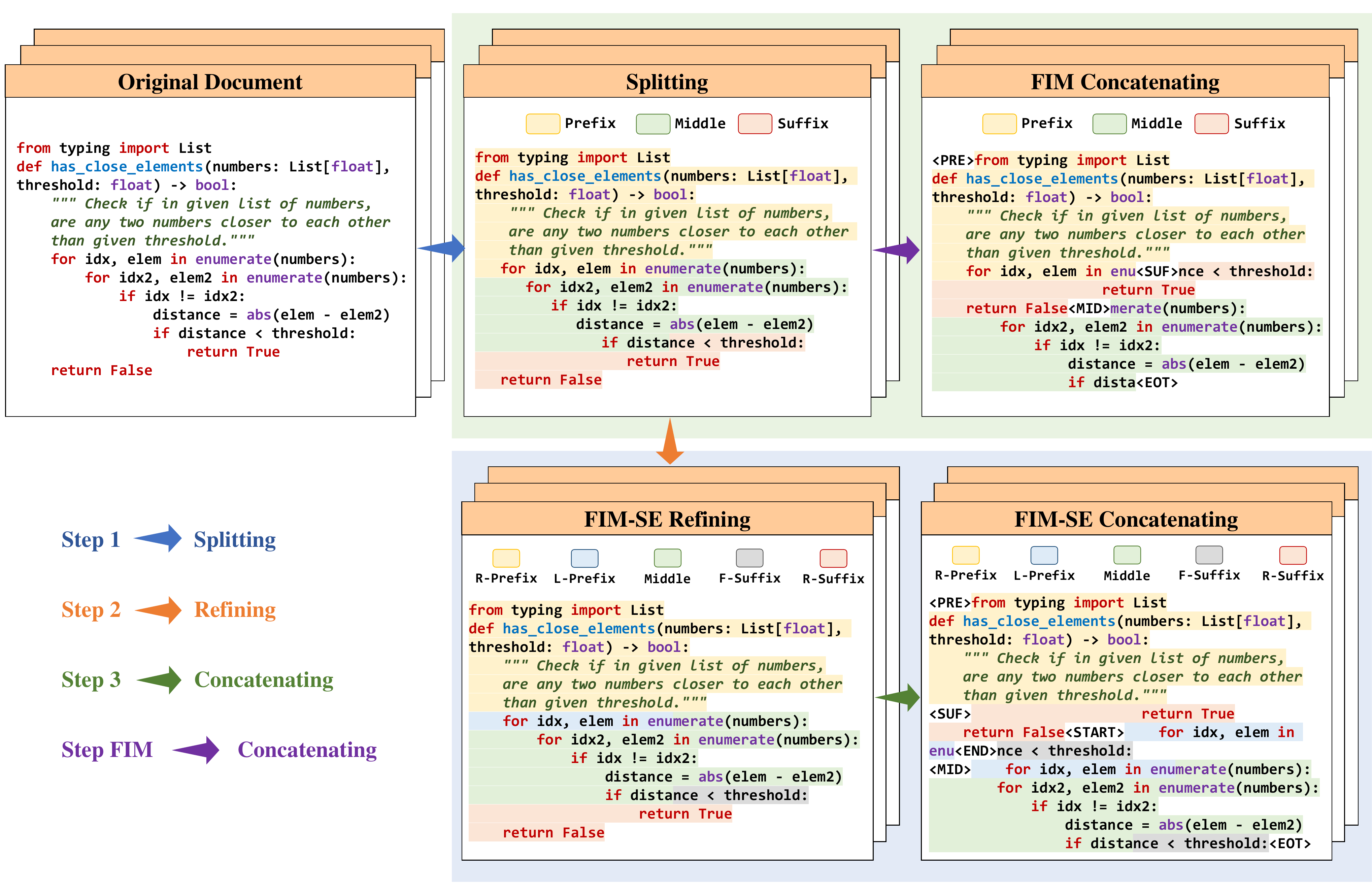}
    \caption{An overview of the difference between FIM and the proposed \model. Here, the green background indicates vanilla FIM and the blue background indicates our \model.}
    \label{fig:method}
\end{figure*}

\section{Methodology} \label{sec:method}

In this section, we introduce the proposed method. 
We begin by outlining the training process with \model, followed by an explanation of the inference procedure. 
Finally, we delve into more training details and highlight the distinctions between our approach and the traditional FIM method.

\subsection{\model Training}

The core idea of \model is to ensure that the tokens predicted by the model are complete, thereby circumventing the issue of large perplexity associated with sub-tokens. Specifically, we shift from character-level to line-level random splitting in training data construction and then reconstruct the prompt to keep the ability of the model on the character-level infilling tasks.

As shown in Figure~\ref{fig:method}, our process for forming the final training sample from a specific document involves three distinct steps.
(1) Splitting: we split the original document into three pieces at the character level, namely the prefix, the middle, and the suffix
(2) Refining: we distinguish between the last line of the prefix and the first line of the suffix, denoting them as \emph{L-Prefix} and \emph{F-Suffix}, respectively. Correspondingly, we label the remaining lines of the prefix and the suffix as \emph{R-Prefix} and \emph{R-Suffix}.
(3) Concatenating: we concatenate all these sections in the following order along with their special tokens as
\begin{equation*}
    \begin{aligned}
        &\text{<PRE>} ~\emph{R-Prefix}~ \text{<SUF>} ~\emph{R-Suffix} \\
        &\text{<START>} ~\emph{L-Prefix}~ \text{<END>} ~\emph{F-Suffix} \\
        &\text{<MID>} ~\emph{L-Prefix}~\emph{Middle}~\emph{F-Suffix}~ \text{<EOT>}.
    \end{aligned}
\end{equation*}

When tokenizing a sample, we tokenize each section individually and then concatenate them with the special tokens, which ensures that special tokens will not be cut or merged. 

\subsection{\model Inference} \label{sec:met_inference}

During the inference stage, the model can be employed for left-to-right generation in a standard manner. When working with an arbitrary location within an existing document, we establish the preceding lines as \emph{R-Prefix} and the following lines as \emph{R-Suffix}. For the specific line at the target location, the text before this point is termed \emph{L-Prefix}, and the text following it is named \emph{F-Suffix}. 
Subsequently, a span is generated to be inserted at this location by autoregressively sampling tokens from the structured prompt
\begin{equation*}
    \begin{aligned}
        &\text{<PRE>} ~\emph{R-Prefix}~ \text{<SUF>} ~\emph{R-Suffix} \\
        &\text{<START>} ~\emph{L-Prefix}~ \text{<END>} ~\emph{F-Suffix} \text{<MID>}.
    \end{aligned}
\end{equation*}
This process continues until the ``<EOT>''~(End of Text) token is produced.

After obtaining the generation, we verify if it begins with the \emph{L-Prefix} and ends with the \emph{F-Suffix}. If the generation does not adhere to these criteria, we classify the infilling endeavor as unsuccessful. Conversely, if the criteria are met, we eliminate the \emph{L-Prefix} from the beginning and the \emph{F-Suffix} from the end, considering the remaining text as the completed segment.

\subsection{Learning and Discussion}

\paratitle{Training Details.} We train our models using the StarCoder code corpus\footnote{\url{https://huggingface.co/datasets/bigcode/starcoderdata}}, a carefully curated dataset sourced from The Stack, encompassing 92 languages. To ensure consistency, we exclude categories such as GitHub issues, GitHub commits, and Jupyter Notebooks, which possess distinct column structures. Additionally, we remove flags marking repositories, files, and stars to maintain a focus on the pure code content in the remaining files.
After gathering the data, we process it using the previously described method with a 90\% FIM rate, following the methodologies of existing studies~\cite{fim2023mohammad,santacoder2023loubna,codellama2023roziere}. It's important to note that we exclusively employ the PSM format depicted in Figure~\ref{fig:method}, as the SPM variant used in prior research~\cite{fim2023mohammad} lacks a separator between the prefix and middle, potentially leading to model confusion. We conduct experiments and give an experimental analysis in Section~\ref{sec:exp_spm}.

\paratitle{Discussion.} Compared to previous masked language modeling on encoder-only models and encoder-decoder models, our method excels in character-level infilling. While these traditional methods primarily concentrate on token-level infilling, this approach often falls short in numerous industry applications, as user text seldom forms complete tokens.
Compared to vanilla FIM~\cite{fim2023mohammad}, our method also has the following two merits. 
Firstly, our method ensures that tokens following ``<MID>'' are complete, eliminating the need for sub-token predictions during inference and thereby mitigating the effects of the large perplexity of the model on sub-tokens.
Secondly, our method transforms character-level infilling into line-level infilling. This unification of formats enhances transfer across different levels, significantly augmenting the efficacy of FIM training.

\begin{table*}[t] \small
\centering
\begin{tabular}{c|c|c|ccc|cc} \toprule
    Model & Size & Training Methods & random-span & single-line & multi-line & Humaneval & MBPP \\ \midrule
    InCoder & 6B & Causal Masking & - & 69.0 & 38.6 & 15.0 & 19.4 \\ 
    FIM & 7B & FIM-SPM & 55.1 & 75.1 & 44.1 & - & - \\ 
    code-davinci-002 & 175B & - & 74.2 & 91.6 & 69.9 & 44.5 & 55.4 \\ \midrule
    \multirow{4.5}{*}{StarCoder} & \multirow{2}{*}{1B} & FIM-PSM & 44.1$^{\ast}$ & 64.3$^{\ast}$ & 30.8$^{\ast}$ & 15.2 & 22.6$^{\ast}$ \\
    ~ & ~ & FIM-SE-PSM & \textbf{48.8~(+4.7)} & \textbf{72.6 (+8.3)} & \textbf{37.1 (+6.3)} & \textbf{16.5} & \textbf{25.6} \\ \cmidrule(lr){2-8}
    ~ & \multirow{2}{*}{15B} & FIM-PSM & 66.4$^{\ast}$ & 83.8$^{\ast}$ & 53.7$^{\ast}$ & 30.4 & 43.2$^{\ast}$ \\
    ~ & ~ & FIM-SE-PSM & \textbf{67.7~(+1.3)} & \textbf{85.8 (+2.0)} & \textbf{57.4 (+3.7)} & \textbf{30.5} & \textbf{44.6} \\ \midrule
    \multirow{6.5}{*}{Code Llama} & \multirow{3}{*}{7B} & FIM-SPM & 39.0 & 83.3 & 50.8 & \multirow{2}{*}{\textbf{33.5}} & \multirow{2}{*}{\textbf{41.4}} \\  
    ~ & ~ & FIM-PSM & 59.7 & 74.1 & 48.2 \\
    ~ & ~ & FIM-SE-PSM & \textbf{67.8~(+8.1)} & \textbf{84.9~(+10.8)} & \textbf{57.2~(+9.0)} & 30.5 & \textbf{41.4} \\ \cmidrule(lr){2-8}
    ~ & \multirow{3.5}{*}{13B} & FIM-SPM & 41.9 & 85.6 & 56.1 & \multirow{2}{*}{36.0} & \multirow{2}{*}{47.0} \\
    ~ & ~ & FIM-PSM & 63.6 & 75.9 & 51.0 \\
    ~ & ~ & FIM-SE-PSM & \textbf{72.4~(+8.8)} & \textbf{87.4~(+11.5)} & \textbf{61.7~(+10.7)} & \textbf{37.2} & \textbf{50.2} \\ \midrule
\end{tabular}
\caption{Pass@1 accuracy on Humaneval infilling datasets. Results evaluated on our end are marked with ``*'', while those unavailable are left blank. Note that StarCoder was evaluated using a cleaned and smaller version of MBPP so we conducted a re-evaluation.}
\label{tab:result}
\end{table*}

\section{Experiments} \label{sec:exp}

In this section, we construct experiments to demonstrate the effectiveness of our method. Due to space limitations, we have constructed more experiments in Appendix~\ref{appendix:exp}.

\subsection{Experimental Setup}

\paratitle{Datasets.} 
Following FIM~\cite{fim2023mohammad}, we use code to test our methods. Because we can use test suites to evaluate the correctness of samples in our tasks even when evaluating long samples from open-ended generations. Specifically, we use three levels of infilling benchmarks, namely random-span, single-line, and multi-line. All of them are constructed from Humaneval benchmarks~\cite{codex2021mark}. Since other infilling benchmarks such as Return Type Prediction and Docstring Generation focus on token-level infilling, we do not use these benchmarks.

\paratitle{Implementation Details.} 
We continually pre-train four models with our methods, \ie StarCoder-1B, StarCoder-15B~\cite{starcoder2023li}, Code Llama 7B, and Code Llama 13B~\cite{codellama2023roziere}. We employ AdamW~\cite{adamw2019ilya} optimizer with $\beta_1 = 0.9$, $\beta_2 = 0.95$, $\epsilon = 10^{-8}$ and weight decay of 0.1. Following previous study~\cite{continue2023kshitij}, we set the peak learning rate to $3 \times 10 ^{-5}$ and use a cosine schedule without warm-up. We use a batch size of 4M tokens which are presented as sequences of 8K tokens each for StarCoder and 16K tokens each for Code Llama. We train each model on 20B tokens in total.
To efficiently train the computationally intensive models, we simultaneously employ DeepSpeed~\citep{rajbhandari2020zero} and Flash Attention 2~\citep{flashattention2tri}. On 32 NVIDIA A800 80GB GPUs, StarCoder-1B, StarCoder-15B, Code Llama 7B, and Code Llama 13B take 14 hours, 140 hours, 75 hours, and 138 hours, respectively.

\subsection{Results}

\paratitle{Baselines.} We compare \model with previous state-of-the-art methods, including InCoder~\cite{incoder2023daniel}, FIM~\cite{fim2023mohammad}, Codex~\cite{codex2021mark}, StarCoder~\cite{starcoder2023li} and Code Llama~\cite{codellama2023roziere}. For other code models such as CodeGeeX~\cite{codegeex2023qinkai} and OctoCoder~\cite{octopack2024niklas}, we do not use them as baselines since they have not undergone infilling pre-training. Since we focus on character-level infilling, models focused on token-level infilling also do not be considered baselines. Because these models cannot effectively handle the character-level infilling task~\cite{fim2023mohammad}.

\paratitle{Random-span.} As shown in Table~\ref{tab:result}, our proposed method demonstrates notable improvements in random-span infilling tasks across four models, specifically achieving gains of 4.7\%, 1.3\%, 8.1\%, and 8.8\%. 
Notably, the enhancement in StarCoder-15B is comparatively modest. This could be attributed to the fact that StarCoder has undergone pre-training with four epochs on a total of 1TB tokens, in contrast to Code Llama's pre-training on 500B tokens, resulting in a more refined model fit. 
Comparing StarCoder-15B with StarCoder-1B, the small model trained on the same tokens has more gain, suggesting that the consistent training approach of our method is particularly beneficial for smaller models in achieving better fit.
Comparing StarCoder-15B with Code Llama 13B, the model with a similar size using fewer tokens achieves better results. This indicates that the consistent training approach of our method accelerates the fitting process in larger models.

\paratitle{Single-line and Multi-line.} The proposed method demonstrates notable improvements in both single-line and multi-line infilling tasks. For instance, based on Code Llama 13B, our method surpasses FIM by 11.5\% and 10.7\% in single-line and multi-line infilling tasks, respectively. This enhancement can be attributed to two key factors. Firstly, our method integrates character-level and line-level processing, significantly enhancing the model's line-level infilling capabilities. Secondly, it avoids the inclusion of any sub-tokens after the ``<MID>'' token, which sharpens the model's accuracy in predicting the initial token. In contrast, in the standard FIM, the first token following ``<MID>'' is typically a sub-token during training, while the model is adopted to predict a complete token in the line-level infilling tasks during the inference stage. A comprehensive case study is provided in Appendix~\ref{appendix:case} for further illustration.

\paratitle{Code Generation Task.} We also report results on  Humaneval~\cite{codex2021mark} and MBPP~\cite{mbpp2021jacob}. As shown in Table~\ref{tab:result}, our method has minimal impact on the model's performance in the two code generation tasks~(Note that we cannot reproduce the result of Code Llama 7B on Humaneval, just 29.9\% in our environment). In summary, FIM-SE demonstrates a remarkable ability to improve infilling tasks without compromising code generation capabilities.

\subsection{Detail Analysis}

\begin{figure}[t]
    \centering
    \includegraphics[width=\columnwidth, page=1]{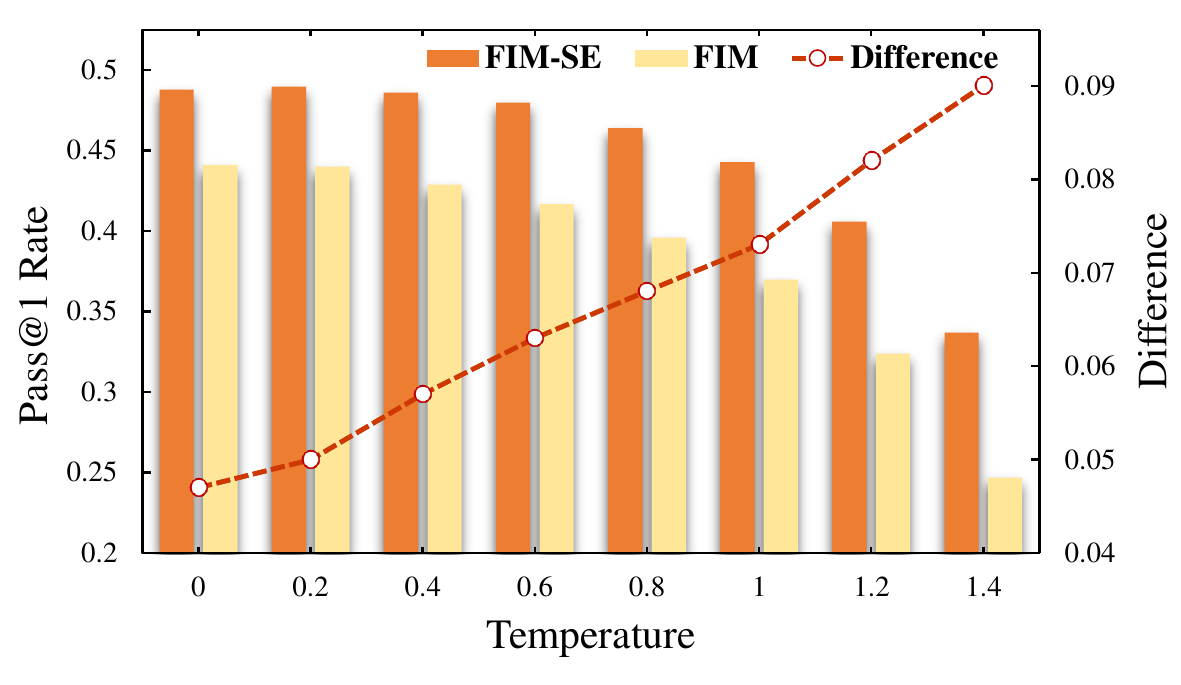}
    \caption{Performance on Humaneval random-span infilling task with different temperatures. The line denotes the difference between \model and FIM. Note that when the temperature surpasses 1.4, both models output noisy text and show very low performance.}
    \label{fig:inconsistent}
\end{figure}

\paratitle{Impact of Inconsistent Labels.} 
As mentioned in Section~\ref{sec:intro} and Section~\ref{sec:met_impact}, we discussed how FIM leads to inconsistent labels during training at split points. This phenomenon results in large perplexity on sub-tokens, subsequently diminishing the model's accuracy in generating sub-tokens. To investigate this effect, we conducted an experiment based on the StarCoder-1B. 
Specifically, we adjusted the temperature within the range of [0, 1.4] and compared the performance of models trained using both \model and FIM in generating 20 completions to estimate the Pass@1 rate. 
Figure~\ref{fig:inconsistent} illustrates that the performance gap between the \model and FIM generators widens as the temperature increases, highlighting the larger perplexity associated with models trained using FIM.

\begin{table}[t] \small
\centering
\begin{tabular}{l|cc} \toprule
    Methods & \model & w/o LF-Loss \\ \midrule
    random-span & \textbf{0.488} & \textbf{0.488} \\ 
    single-line & \textbf{0.726} & 0.716 \\ 
    multi-line & \textbf{0.371} & 0.369 \\ 
    Test loss & 0.847 & \textbf{0.834} \\ \bottomrule
\end{tabular}
\caption{Effect of training loss on sub-tokens. The metric is Pass@1 accuracy. Here, LF-Loss denotes the loss for tokens in \emph{L-Prefix} and \emph{F-Suffix}.}
\label{tab:inconsistent}
\end{table}

Furthermore, we evaluate the impact of inconsistent labels on training. Specifically, we mask the loss for tokens in \emph{L-Prefix} and \emph{F-Suffix}, ensuring that only complete tokens contribute to loss calculations. As shown in table~\ref{tab:inconsistent}, computing losses for \emph{L-Prefix} and \emph{F-Suffix} led to a slightly higher test loss without significantly affecting performance. This could be attributed to the minimal proportion of sub-tokens, as the presence of up to four sub-tokens per sample had a negligible impact on the final test results. In summary, while the loss of sub-tokens in training scarcely affects performance, the presence of sub-tokens in prediction objectives markedly influences performance.

\begin{table}[t] \small
\centering
\begin{tabular}{c|ccc} \toprule
    Methods & random-span & single-line & multi-line \\ \midrule
    \model & 0.488 & \textbf{0.726} & 0.371 \\ 
    SPM v1 & \textbf{0.492} & 0.703 & 0.374 \\
    SPM v2 & 0.013 & 0.085 & 0.088 \\
    SPM v3 & 0.090 & 0.717 & \textbf{0.383} \\ \bottomrule
\end{tabular}
\caption{Comparison between different SPM format variants and FIM-SE. The metric is Pass@1 accuracy.}
\label{tab:spm}
\end{table}

\paratitle{Comparison with SPM Mode.} \label{sec:exp_spm}
In previous studies, the Suffix-Prefix-Middle variant had better performance in most cases~\cite{fim2023mohammad,codellama2023roziere}. Here, we explore how to combine our method with SPM mode based on StarCoder-1B. Specifically, we designed the following three prompt formats for SPM mode. We train the model using these formats and the PSM mode, equally distributed across 20 billion tokens.
\begin{enumerate}[label=(\arabic*), itemsep=1pt]
    \item \uline{SPM v1}: ``<SUF> \emph{R-Suffix} <PRE> \emph{R-Prefix} <START> \emph{L-Prefix} <END> \emph{F-Suffix} <MID>'', which add the constraints before the middle to the vanilla SPM mode.
    \item \uline{SPM v2}: ``<PRE> <SUF> \emph{R-Suffix} <START> \emph{L-Prefix} <END> \emph{F-Suffix} <MID> \emph{R-Prefix}'', which add the constraints before the middle to the variant SPM in FIM.
    \item \uline{SPM v3}: ``<PRE> <SUF> \emph{R-Suffix} <MID> \emph{R-Prefix} <START> \emph{L-Prefix} <END> \emph{F-Suffix}'', which add the constraints after prefix to the variant SPM in FIM.
\end{enumerate}

Table~\ref{tab:spm} presents all comparison results of the three variants. As we can see, \uline{SPM v2} and \uline{SPM v3} perform worse on random-span infilling tasks. This occurs because there is no separator between the prefix and the middle, leading to conflicts with the PSM mode, regardless of where the restriction is inserted. In contrast, \uline{SPM v1} and PSM perform almost the same because there is no conflict. To maintain consistency with the pre-trained models~\cite{starcoder2023li,codellama2023roziere}, we adopt the PSM mode.

\begin{figure}[t]
    \centering
    \subfigure[PCP Rate of \model.]{
    \includegraphics[width=\columnwidth, page=2]{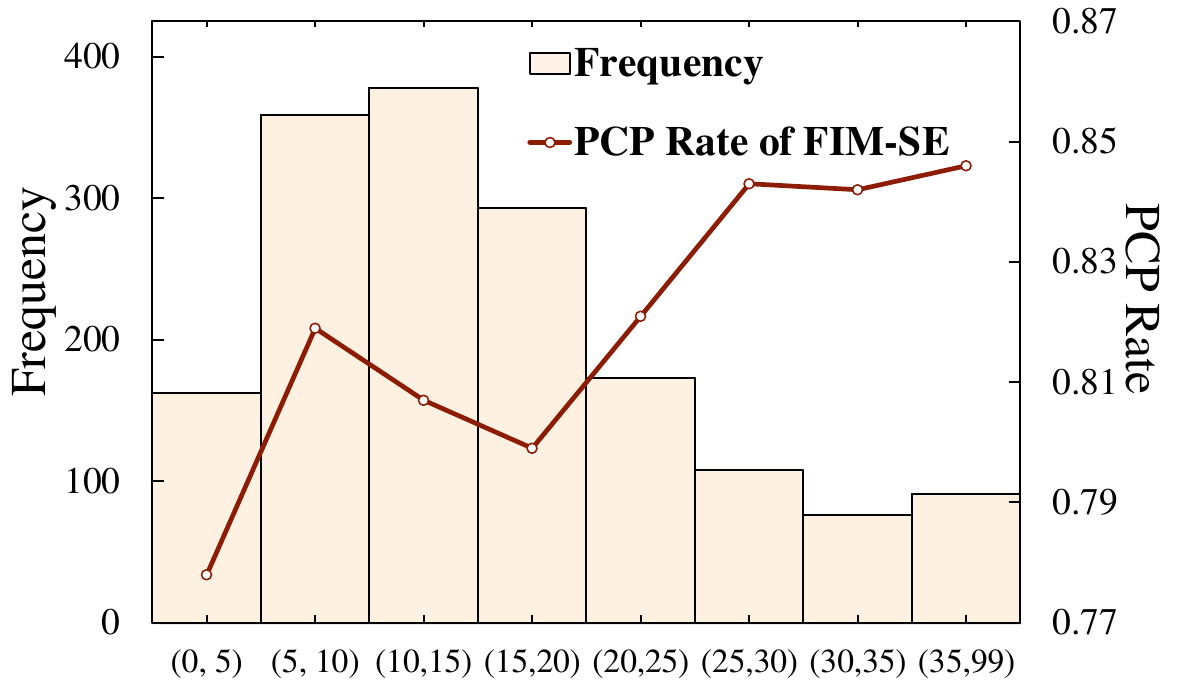}
    }
    \subfigure[Pass@1 performance of \model and FIM.]{
    \includegraphics[width=\columnwidth, page=3]{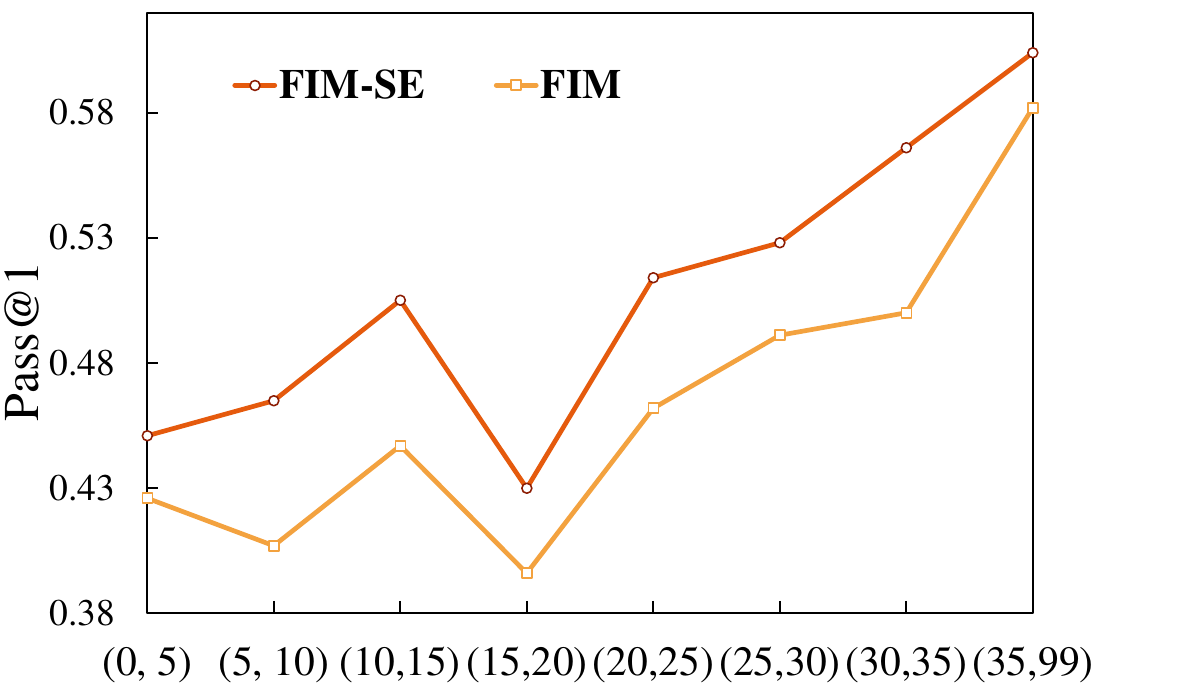}
    }
    \caption{Statistics of length of \emph{L-Prefix} and \emph{F-Suffix}.}
    \label{fig:length}
\end{figure}

\paratitle{Analysis of Post-Check during Inference.}
As we mentioned in Section~\ref{sec:met_inference}, it's essential to verify if the generation begins with the \emph{L-Prefix} and ends with the \emph{F-Suffix}. 
Here, we perform statistical analysis on the success rate of the model based on StarCoder-1B. We focus on the Post-Check Pass Rate (PCP Rate), which quantifies the percentage of model outputs complying with the post-check criteria, \ie starting with the \emph{L-Prefix} and ending with the \emph{F-Suffix}. We then examine the correlation between the PCP Rate and the average length of the \emph{L-Prefix} and \emph{F-Suffix}. Additionally, we analyze the Pass@1 rates for FIM-SE and FIM across varying lengths of these prefixes and suffixes.

As shown in Figure~\ref{fig:length}, the PCP Rate increases with length, suggesting that longer \emph{L-Prefixes} and \emph{F-Suffixes} provide more guidance for the model's text completion. Moreover, the Pass@1 metrics for both \model and FIM also support this, showing enhanced performance with extended \emph{L-Prefixes} and \emph{F-Suffixes}. Across all lengths, \model consistently outperforms the standard FIM, demonstrating the effectiveness of our approach.
\section{Conclusion}

In this paper, we showed that traditional infilling techniques struggle with managing the boundaries of prefixes and suffixes. To address this, we introduced a novel approach, referred to as \model. Our method transforms the random-span mode to multi-line mode by removing the \emph{L-Prefix} and \emph{F-Suffix}. We further incorporated two special tokens to delineate the two incomplete lines, thereby guiding the generation. Extensive experiments reveal that our approach surpasses existing baselines with a clear edge. In future work, we plan to explore the adaptation of our method to the variant SPM mode, which holds the promise of even better performance.

\section{Limitations}

The primary limitation of this study is the inability of the proposed method to accommodate the variant SPM mode, previously established as superior by prior research. This challenge arises due to the absence of a distinct delimiter between the prefix and middle, impeding our capacity to guide the model on the commencement point for completion and to appropriately position the prompt that instructs the model to start with \emph{L-Prefix} and end with \emph{F-Suffix}. In future endeavors, we plan to explore adapting our method to the variant SPM mode, to achieve better performance.
Another limitation of this paper is the probability that our proposed method fails to complete tasks when the generation neither starts with the \emph{L-Prefix} nor ends with the \emph{F-Suffix}. For example, the fail rates of StarCoder-1B and StarCoder-15B are 18.7\% and 9.4\%, respectively. This issue is a primary factor impacting model performance. Future work will concentrate on improving the post-check pass rate by developing more comprehensive prompts and refining constraint decoding.

\section{Ethics Statement}

In this paper, we utilized the StarCoder dataset~\cite{starcoder2023li}. This dataset has been made publicly available for academic purposes. The creators of the StarCoder dataset have transparently disclosed its derivation from The Stack v1.2~\cite{stack2022denis}. Importantly, The Stack v1.2 is compiled from a collection of GitHub repositories, all of which operate under permissive licenses. This ensures that the data's utilization aligns with the original authors' intentions and the legal frameworks governing open-source contributions.
In conclusion, the application of the StarCoder dataset in our study complies with the ethical guidelines for research data usage, aligning with the broader principles of academic honesty and the responsible conduct of research.

\section*{Acknowledgment}
This project is funded in part by National Key R\&D Program of China Project 2022ZD0161100, by the Centre for Perceptual and Interactive Intelligence (CPII) Ltd under the Innovation and Technology Commission (ITC)’s InnoHK, by General Research Fund of Hong Kong RGC Project 14204021. Hongsheng Li is a PI of CPII under the InnoHK. 

\bibliography{references}

\clearpage\appendix\section*{Appendix}

\section{Fill-In-the-Middle~(FIM)} \label{appendix:fim}

In Section~\ref{sec:pre_fim}, We briefly introduced the prefix-suffix-middle~(PSM) mode of FIM~\cite{fim2023mohammad}. Here, we give a detailed description of the suffix-prefix-middle~(SPM) mode and a variant SPM mode.

For the vanilla SPM mode, it just swaps the prefix and the suffix. Specifically, after splitting, it moves the suffix to the before:
\begin{equation*}
    \text{doc} \to \text{(pre, mid, suf)} \to \text{(suf, pre, mid)},
\end{equation*}
then concatenate the three pieces using special tokens as
\begin{equation*}
    \text{<SUF> suf <PRE> pre <MID> mid <EOT>}.
\end{equation*}

To maximize transfer between PSM mode and SPM mode, FIM proposed a novel variant of SPM mode, which concatenates the prefix, the middle, and the suffix pieces as
\begin{equation*}
    \text{<PRE> <SUF> suf <MID> pre mid <EOT>}.
\end{equation*}
The format occurs naturally as part of PSM training when the chosen prefix is empty. In this way, the two modes have a consistent format and they can transfer with each other in joint training and maximize the profits.

\section{Additional Experiments} \label{appendix:exp}

\subsection{Comparison with Token Healing} \label{appendix:healing}

As discussed in Section~\ref{sec:rel}, token healing is proposed as an ideal solution for addressing tokenization that normally arise at the boundary between the end of a prompt and the beginning of a set of generated tokens. Here, we evaluate our approach against the token healing method based on StarCoder-1B. However, since token healing struggles with the split points at the end of generated tokens and the subsequent suffix, we integrate it with our method for a comprehensive solution. Specifically, we construct the prompt as ``<PRE> \emph{R-Prefix} <SUF> \emph{R-Suffix} <START> \emph{L-Prefix} <END> \emph{F-Suffix} <MID> \emph{L-Prefix}'' and focus solely on verifying if the generated text ends with \emph{F-Suffix}.

\begin{table}[t] \small
\centering
\begin{tabular}{c|c} \toprule
    Methods & random-span \\ \midrule
    \model & 0.488  \\ 
    Token Healing & 0.484 \\ \bottomrule
\end{tabular}
\caption{Comparison with Token Healing.}
\label{tab:healing}
\end{table}

\begin{table}[t] \small
    \centering
    \subtable[A case can be solved by token healing.]{
    \centering
    \label{tab:healing_right}
    \begin{tabular}{c|c|ccccc} \toprule
        Piece & Raw Text & \multicolumn{5}{c}{Tokens} \\ \midrule
        \multirow{1}{*}{Prefix} & \multirow{1}{*}{\makecell[l]{def \textcolor{red}{so}}}
        & def & \_~\textcolor{red}{so} \\ \midrule
        \multirow{1}{*}{Output} & \multirow{1}{*}{\makecell[l]{def \textcolor{red}{sort}(arr)}}
        & def & \_~\textcolor{red}{sort} & ( & arr & ) \\ \midrule
        \multirow{1}{*}{Label} & \multirow{1}{*}{\makecell[l]{def \textcolor{red}{sort}(arr)}}
        & def & \_~\textcolor{red}{sort} & ( & arr & ) \\ \bottomrule
    \end{tabular}
    }
    \subtable[A case cannot be solved by token healing.]{
    \centering
    \label{tab:healing_wrong}
    \begin{tabular}{c|c|cccc} \toprule
        Piece & Raw Text & \multicolumn{4}{c}{Tokens} \\ \midrule
        \multirow{2}{*}{Prefix} & \multirow{2}{*}{\makecell[l]{r.add(\textcolor{red}{delim}}}
        & r & . & add & ( \\
        ~ & ~ & \textcolor{red}{delim} \\ \midrule
        \multirow{2}{*}{Output} & \multirow{2}{*}{\makecell[l]{r.add(\textcolor{red}{delimter})}}
        & r & . & add & (  \\ 
        ~ & ~ & \textcolor{red}{delim} & \textcolor{red}{ter} & ) \\ \midrule
        \multirow{2}{*}{Label} & \multirow{2}{*}{\makecell[l]{r.add(\textcolor{red}{delimeter})}}
        & r & . & add & (  \\ 
        ~ & ~ & \textcolor{red}{deli} & \textcolor{red}{meter} & ) \\ \bottomrule
    \end{tabular}
    }
    \caption{Case of token healing. The first case can be perfectly solved by token healing. The second case cannot be solved by token healing. Here, `\_' denotes blank.}
\end{table}

Table~\ref{tab:healing} presents the comparison results between our method and token healing. 
Surprisingly, token healing performs slightly worse than our method. 
Detailed analysis revealed that token healing struggles with complex scenarios, such as splitting the last token into two sub-tokens and merging the latter sub-token with the initial generated token.

\begin{table*}[t] \small
    \centering
    \renewcommand{\arraystretch}{1.1}
    \begin{tabular}{l} \toprule
    \textbf{Prefix} \\ \midrule
     \begin{lstlisting}[language=Python]
def largest_prime_factor(n: int):
    """Return the largest prime factor of n. Assume n > 1 and is not a prime.
    >>> largest_prime_factor(13195)
    29
    """
    def is_prime(k):
        if k < 2:
            return False
        for i in range(2, k - 1):
            if k % i == 0:
                return False
        return True           
    \end{lstlisting} \\ \midrule
    \textbf{Suffix} \\ \midrule
    \begin{lstlisting}[language=Python]
    for j in range(2, n + 1):
        if n % j == 0 and is_prime(j):
            largest = max(largest, j)
    return largest
    \end{lstlisting} \\ \midrule
    \textbf{Target Middle} \\ \midrule
    \begin{lstlisting}[language=Python]
    largest = 1
    \end{lstlisting} \\  \midrule
    \textbf{The top five choices for the initial generated token on StarCoder-1B~(FIM), along with their probabilities} \\ \midrule
    `\texttt{\textbackslash}n': 0.463; `\texttt{\textbackslash}n\_~\_~\_~': 0.225; `\_~\_~\_~\_': 0.073; `<|endoftext|>': 0.068; `\_~\_~\_~\_\texttt{\textbackslash}n\_~\_~\_~': 0.061; \\ \midrule
    \textbf{The top five choices for the initial generated token on StarCoder-1B~(\model), along with their probabilities} \\ \midrule
    `\_~\_~\_~': 0.829; `\texttt{\textbackslash}n\_~\_~\_~': 0.115; `\_~\_~\_~\_\texttt{\textbackslash}n\_~\_~\_~': 0.037; `\_~\_~\_~\_': 0.008; `\_~\_~\_~\_~\_~\_~\_~\_\texttt{\textbackslash}n\_~\_~\_~' : 0.002; \\ \bottomrule
    \end{tabular}
    \caption{A Case to show perplexity of models on the initial generated token. Here, `\_' denotes blank, and `\texttt{\textbackslash}n' denotes newline.}
    \label{tab:case}
\end{table*}

Here, we provide an in-depth analysis of the situation. Token healing backs up the generation process by one or more tokens before the end of the prompt, then constrains the first tokens generated to have a prefix that matches the last token in the prompt. As illustrated in Table~\ref{tab:healing_right}, if the last token in the prompt is "so," token healing identifies a token that both matches this last token's prefix and possesses the highest probability, such as "sort." Consequently, the first generated token is seamlessly integrated, allowing the generation process to proceed smoothly.

However, due to the consistent integrated intrinsic features of the SentencePiece~\cite{sentencepiece2018taku} algorithm, it tends to merge the last sub-token with the preceding one if possible. For example, as illustrated in Table~\ref{tab:healing_wrong}, the word "delimiter" is tokenized into "deli" and "meter." If a prompt ends with "delim", the algorithm prefers tokenizing this as "delim" instead of splitting it into "deli" and "m". Token healing does not intervene, because there is no token starting with "delim". Consequently, when the last sub-token can be combined with the previous one, token healing is unable to rectify it effectively.

To effectively resolve this issue, it is necessary to revert several tokens and subsequently engage in limited decoding, utilizing a Trie tree, until the regenerated text encompasses the previously rolled-back tokens. Nonetheless, this approach is time-intensive as each decoding step requires traversing the Trie tree to identify all tokens corresponding to the given prefix. In contrast, our method only requires modifying the prompt and doing some post-processing. In addition, our method can handle both boundaries between the prefix and middle as well as boundaries between the suffix and middle.

\subsection{Case Study} \label{appendix:case}

Here, we present a case demonstrating the model's large perplexity on the initial generated token based on StarCoder-1B. Table~\ref{tab:case} illustrates that, despite being a single-line infilling scenario, the perplexity for the first token is remarkably large, significantly influencing the overall generation. This is primarily because the first token following ``<MID>'' tends to be a sub-token during training, varying across samples due to random character-level splitting. In contrast, our approach guarantees that no sub-token prediction is required, leading to lower perplexity and enhanced performance.

Based on these concerns, we hypothesize that the superior performance of the variant SPM mode over the PSM mode, particularly evident in the single-line infilling task on Code Llama~\cite{codellama2023roziere} (85.6\% vs. 75.9\%), can be attributed to the specific processing format employed by Code Llama. In the format ``<PRE> <SUF> suf <MID> pre mid <EOT>'', Code Llama initially merges the prefix and middle segments before tokenization. This approach ensures that, following the ``<MID>'' token, there are no sub-tokens except for the initial token. Consequently, this format also guarantees that no sub-token prediction is required when the prefix is not empty, contributing to the enhanced performance of the variant SPM mode.
In contrast, FIM~\cite{fim2023mohammad} adopts a different approach by tokenizing the prefix and the middle separately before concatenating them. This leads to the presence of sub-tokens amidst the tokens. Consequently, the performance gap between SPM and PSM modes in FIM is narrower (61.6\% vs. 62.2\%) compared to that in Code Llama.

\end{document}